\documentclass[10pt]{article} 
\usepackage[preprint]{tmlr}

\usepackage{graphicx}
\usepackage{algorithm}
\usepackage{algorithmic}

\usepackage{amsfonts}
\usepackage{booktabs}
\usepackage{caption}

\usepackage{amsmath,amsfonts,bm}

\def\eqref#1{equation~\ref{#1}}

\def\1{\bm{1}}

\DeclareMathAlphabet{\mathsfit}{\encodingdefault}{\sfdefault}{m}{sl}
\SetMathAlphabet{\mathsfit}{bold}{\encodingdefault}{\sfdefault}{bx}{n}

\usepackage{hyperref}
\usepackage{url}

\usepackage{cleveref}

\title{One Model, Many Depths: \\Straightening Latent Trajectories \\for Post-Hoc Depth Reduction}

\author{\name Siddharth Rout \email siddharth.rout@ubc.ca \\
      \addr University of British Columbia
      \ANDH
      \name Nguyen Thi Hong Linh \email nguyen.linh11@northeastern.edu \\
      \addr Northeastern University
      \ANDH
      \name Peter Yichen Chen \email peter.chen@ubc.ca\\
      \addr University of British Columbia
      \ANDH
      \name Eldad Haber \email ehaber@eoas.ubc.ca\\
      \addr University of British Columbia}

\def\month{MM}  
\def\year{2027} 
\def\openreview{\url{https://openreview.net/forum?id=XXXX}} 

\begin{document}

\maketitle

\begin{abstract}
Stacked layers in a transformer can be read as steps of a discretized ODE acting on a latent state. In standard models, this latent trajectory is strongly curved, and pruning layers after training sharply degrades performance. We propose a trajectory-straightening regularizer that penalizes each layer's update for deviating from the average update, which pushes successive layers to move the hidden state in a shared direction. At inference, an extrapolation rule replaces dropped layers by repeating the last computed update. We test the idea on vision transformers (for classification of MNIST, CIFAR100, satellite cloud images) and on diffusion transformers (for MNIST super-resolution, Moving MNIST video prediction, 2m temperature nowcasting). The regularizer lowers a directional-misalignment measure by one to several orders of magnitude and, at moderate strengths, largely preserves full-depth quality on most tasks. Models trained with it degrade far more gracefully than unconstrained models when layers are removed, and more gracefully than LayerDrop under aggressive reductions, although LayerDrop is as good or better for mild reductions. The costs are real: lower full-depth quality at large $\lambda$, steep accuracy loss under aggressive depth reduction on harder tasks, and slower optimization. We compare against LayerDrop, a structured-dropout baseline, on every task. Our evidence is that shaping the geometry of latent trajectories is a promising route to depth-flexible models, not that it already beats existing alternatives.
\end{abstract}

\section{Introduction}

Large neural networks, including large language models (LLMs) and vision--language models (VLMs), lead many domains \cite{vaswani2017attention, devlin2019bert, touvron2023llama, achiam2023gpt, gu2024mamba, dosovitskiy2020image, liu2021swin, peebles2023scalable, zhu2024vision}. Their hidden representations evolve through long chains of nonlinear transformations that remain poorly understood. That chain is also the main inference cost: every input must pass through every layer.

A deep network can be read as the discretization of a high-dimensional ordinary differential equation (ODE), in which the output of each layer is one step in latent space \cite{haber2017stable}. The hidden states then form a trajectory in latent space. In general this trajectory is highly curved, so each layer performs a distinct transformation, and removing layers after training degrades predictions sharply. Also, hence, pruning a noticeable number of hidden layers is not trivial \cite{fan2019reducing}.

This raises a natural question: \emph{can the trajectory be made approximately straight, and does that make depth reduction at inference less harmful?} If successive layers update the state in a common direction, the layers we drop can be approximated by repeating one computed update. In the idealized limit, the cost of an $N$-layer network falls by a factor approaching $N$. We treat this as a limit, not an achieved result: how closely it can be approached, and at what cost in accuracy, is the empirical question we study.

We add a \emph{trajectory-straightening regularizer} to the training loss that penalizes disagreement between the updates of different layers. At inference, a simple extrapolation rule replaces the dropped layers by repeating the last computed update. This differs from post-hoc layer pruning \cite{gromov2025unreasonable, men2025shortgpt}, which acts on a finished network, and from structured dropout \cite{fan2019reducing}, which drops layers at random during training. Our approach shapes the geometry of the layer, updates themselves.

\paragraph{Contributions.}
\begin{enumerate}
    \item \textbf{A geometric regularizer.} We cast the sequence model as a discrete high-dimensional ODE and add a straightness penalty on its latent trajectory.
    \item \textbf{A depth-reduction rule.} One-step extrapolation replaces dropped layers with the update from the last computed layer.
    \item \textbf{An empirical characterization.} We map the trade-off between regularization strength, number of layers removed, and task quality on prediction and classification tasks, showing where the approach helps and where it does not.
    \item \textbf{Training behavior.} Stronger constraints slow convergence but reach similar minima at moderate strengths, and very large coefficients need a scheduled increase to train stably.
\end{enumerate}

\begin{table*}[!th]
\centering
\caption{Qualitative comparison of depth-flexible approaches. ``Static graph'': the computational graph does not depend on the input. ``Post-hoc depth'': the number of layers can be chosen at inference without retraining.}
\label{tab:related}
\begin{tabular}{lccc}
\hline
Approach & Gate/exit heads & Static graph & Post-hoc depth \\
\hline
Early exit \cite{teerapittayanon2016branchynet, huang2017multi} & Yes & No & Via exit choice \\
Layer skip \cite{wang2018skipnet, veit2018convolutional} & Yes & No & Input-based \\
Mixture-of-Depths \cite{raposo2024mixture} & Yes (routers) & Yes & No \\
Structured dropout \cite{fan2019reducing} & No & Yes & Yes \\
Neural ODE/DEQ \cite{chen2018neural, bai2019deep} & No & Depends & Yes \\
Post-hoc pruning \cite{gromov2025unreasonable, men2025shortgpt} & No & Yes & Yes \\
Trajectory straightening (ours) & No & Yes & Yes \\
\hline
\end{tabular}
\end{table*}

\paragraph{Scope.}
We do not claim that our method matches existing pruning or early-exit approaches in accuracy--cost trade-offs. Apart from LayerDrop \cite{fan2019reducing}, which we include as a baseline on every task, we include no head-to-head comparisons with them. The other comparisons are between models trained with and without the regularizer. We also document the costs: lower full-depth quality at large $\lambda$, sharp degradation under aggressive depth reduction on harder tasks, and slower training.

\paragraph{Why Depth flexibility?}
Practical deployment rarely has a fixed computational budget: hardware, load, energy, and latency constraints vary. A \emph{depth-flexible} model serves several inference depths from one set of parameters, so heterogeneous devices and fluctuating workloads need no separately trained models. This requires intermediate representations to stay useful across depths, which is the property the regularizer targets. Video prediction is one natural setting. Large networks are increasingly used for video understanding \cite{qian2024streaming, shu2025video}, generation \cite{zheng2024open, tian2024videotetris}, editing \cite{li2024vidtome, deng2024dragvideo}, and inverse problems \cite{daras2024warped, kwon2025vision}, and prediction difficulty varies across frames and scenes with their motion characteristics \cite{shi2017deep, wu2021motionrnn}. A model with several depths could, in principle, use a shallow configuration for easy frames and full depth for hard ones \cite{ilhan2024adaptive}. We take a first step: whether one trained model can offer a family of accuracy--latency operating points chosen by the user at inference. Choosing depth automatically from input difficulty is future work.

\section{Related Work}

\paragraph{Adaptive depth.}
Dynamic neural networks \cite{han2021dynamic} vary computation per input by early exiting \cite{huang2017multi, teerapittayanon2016branchynet, bolukbasi2017adaptive} or by skipping layers \cite{graves2016adaptive, wang2018skipnet, veit2018convolutional}, including in transformers \cite{elbayad2019depth, liu2021faster}. These methods rely on gates or exit heads and typically yield irregular graphs that run inefficiently on GPUs. Mixture-of-Depths \cite{raposo2024mixture} restores a static graph through token-level routing, but depth cannot be selected after training. Structured dropout \cite{fan2019reducing} trains for depth reduction by dropping layers at random; we regularize the geometry of the layer updates instead.

\paragraph{ODE views and trajectory straightening.}
Neural ODEs \cite{chen2018neural} and related models \cite{rubanova2019latent, bai2019deep} treat depth as continuous, so the number of solver steps can vary at test time. Kinetic-energy and Jacobian penalties \cite{finlay2020train} make such ODEs cheaper to solve by bounding the magnitude of the dynamics. Rectified flow \cite{liu2022flow} straightens generative trajectories so that few integration steps suffice. The perceptual straightening hypothesis \cite{henaff2019perceptual} holds that visual representations straighten natural video, easing linear extrapolation. We apply a similar idea along the depth axis of a discrete network with untied layers: our penalty couples the updates of different layers so that one update can stand in for several. It belongs to the family of regularizers that control layer-to-layer dynamics \cite{yoshida2017spectral, bansal2018can, rifai2011contractive, arjovsky2017wasserstein} but targets trajectory geometry directly.

\paragraph{Layer redundancy and weight tying.}
Trained networks are often redundant in depth. Residual networks act like ensembles of shallow paths and iteratively refine a shared representation \cite{veit2016residual, greff2016highway}, and the deeper layers of pretrained LLMs can often be removed with little loss \cite{gromov2025unreasonable, men2025shortgpt, sun2025transformer}. Unconstrained baselines may therefore tolerate some layer removal, so, where we have them, we compare against them under the same dropping protocol. Weight-tied models \cite{dehghani2018universal, lan2019albert} repeat a block by construction. We keep untied weights and repeat the last additive \emph{update}, and the regularizer encourages update alignment instead of imposing it architecturally.

\Cref{tab:related} summarizes the differences. We see these approaches as complementary: the regularizer could be combined with an exit head or a learned depth selector. We make no claim about relative accuracy--cost trade-offs. To our knowledge, the combination of (1) a discrete-ODE view of the layers, (2) an explicit layer-update alignment penalty during training, and (3) replacing dropped layers by repeating the last computed update has not been evaluated before, though each ingredient has close relatives above.

\section{Method}

\subsection{Setup}

\begin{figure}[t]
\centering
\includegraphics[width=\textwidth]{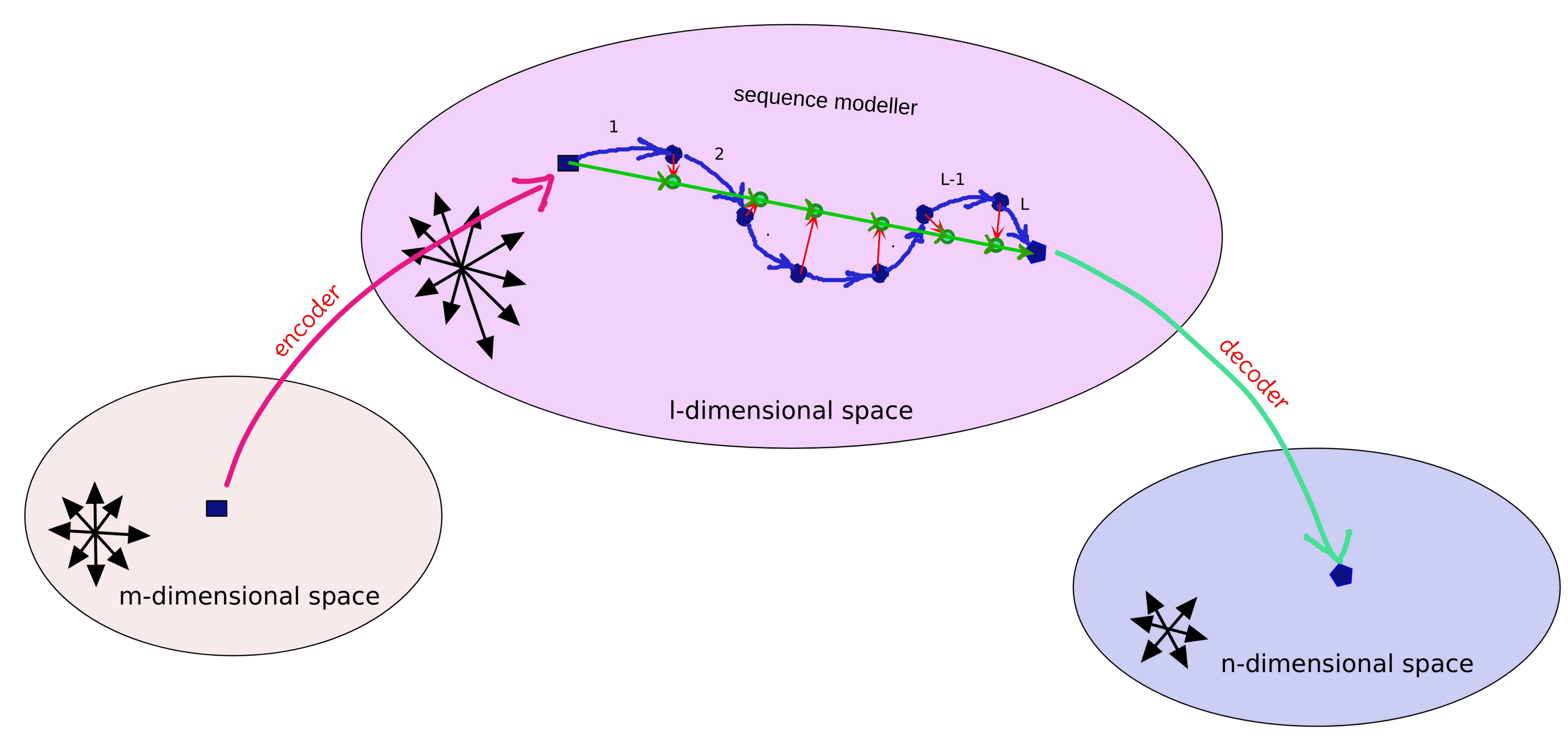}
\caption{Graphical representation of a large model, where the encoder transforms the $m$-dimensional input data to an $l$-dimensional latent space and the decoder transforms the latent representation to $n$-dimensional output. The sequence modeller (in blue) can be seen as an $L$-step operation within the latent space, where $L$ is the number of hidden layers. The sequence modeller can be seen to be straightened (in green) using our regularizer. Straightening makes the per-layer steps along the latent trajectory (small arrows) point in a common direction, so a dropped layer can be replaced by repeating the last computed step.}
\label{fig:largemodel}
\end{figure}

We view a large model as three trainable functions applied in order: an encoder $\mathcal{E}:\mathbb{R}^m\rightarrow\mathbb{R}^l$ that maps an input to a latent state, a sequence modeller $\mathcal{S}$ of $L$ layers that evolves that state, and a decoder $\mathcal{D}:\mathbb{R}^l\rightarrow\mathbb{R}^n$ that maps the final state to the output (\Cref{fig:largemodel}). Each layer $\mathcal{S}_i:\mathbb{R}^l\rightarrow\mathbb{R}^l$ has its own parameters and can be a transformer block, a recurrent block, an SSM block \cite{vaswani2017attention, gu2024mamba}, or similar. Then the neural network is 
\begin{equation}
    \mathcal{N}(x) = \mathcal{D}\big(\mathcal{S}_L\circ\mathcal{S}_{L-1}\circ\cdots\circ\mathcal{S}_1(\mathcal{E}(x))\big).
\end{equation}
The latent state after layer $i$ is $\mathcal{F}_i(x)=\mathcal{S}_i\circ\cdots\circ\mathcal{S}_1\circ\mathcal{E}(x)$, with $\mathcal{F}_0(x)=\mathcal{E}(x)$, and the update made by layer $i$ is $\Delta_i=\mathcal{F}_i(x)-\mathcal{F}_{i-1}(x)$. Read as an ODE discretization, $\mathcal{F}_0,\dots,\mathcal{F}_L$ is the latent trajectory and each $\Delta_i$ is one step along it. The trajectory is straight, with equal steps, exactly when all $\Delta_i$ are identical. For surveys of such models see \citet{ghosh2024exploring, liu2025vision}.

\subsection{Straightening Loss}

We penalize each update's deviation from the mean update $\bar{\Delta}(x)=\frac{1}{L}\big(\mathcal{F}_L(x)-\mathcal{F}_0(x)\big)$:
\begin{equation}
    \mathcal{L}_{\mathrm{straight}} = \frac{1}{|\mathcal{B}|}\sum_{x\in\mathcal{B}}\sum_{j=1}^{L}\big\|\Delta_j(x)-\bar{\Delta}(x)\big\|_2^2,
\end{equation}
\begin{equation}
    \mathcal{L} = \mathcal{L}_{\mathrm{MSE}} + \lambda\,\mathcal{L}_{\mathrm{straight}},
\end{equation}
where $\mathcal{B}$ is a mini-batch, $\mathcal{L}_{\mathrm{MSE}}$ is the data-fitting loss, and $\lambda$ is the straightening coefficient (\Cref{alg:nn_training}). The straightening loss is zero only when every update equals the mean update. We call this quantity the step-consistency loss (step loss for short). It is zero if and only if every update equals the mean update, so it penalizes differences in both direction and step size. Very large $\lambda$ trains unstably and slowly, so for those runs we increase $\lambda$ on a logarithmic schedule.

\subsection{Depth Reduction by Extrapolation}

To drop $N_D$ layers at inference, we run only the first $L-N_D$ layers and repeat the last computed update $N_D$ times in place of the dropped ones (\Cref{alg:nn_inference}). With $a_0=\mathcal{E}(x)$ and $a_l=\mathcal{S}_l(a_{l-1})$ for $l=1,\dots,L-N_D$,
\begin{equation}
    \hat{a} = a_{L-N_D} + N_D\,\big(a_{L-N_D}-a_{L-N_D-1}\big), \qquad \hat{y}=\mathcal{D}(\hat{a}).
\end{equation}
With $N_D=0$ this is the ordinary forward pass. If all updates were identical, the extrapolation would reproduce the full network exactly. In practice its quality depends on how close training gets to that limit. We keep untied weights throughout: the regularizer encourages aligned updates, but the architecture does not impose them.

\begin{algorithm}[H]
\caption{Training a straightened network $\mathcal{N}_\theta$ with $L$ layers}
\label{alg:nn_training}
\begin{algorithmic}[1]
\REQUIRE Training data $\{(x_i, y_i)\}_{i=1}^N$, learning rate $\eta$, epochs $T$, coefficient $\lambda$
\ENSURE Trained parameters $\theta$
\STATE Initialize $\theta$ randomly
\FOR{epoch $= 1$ to $T$}
    \FOR{each mini-batch $\mathcal{B}\subset\{(x_i, y_i)\}_{i=1}^N$}
        \STATE $\hat{y}_i \leftarrow \mathcal{N}_\theta(x_i)$ for all $(x_i, y_i)\in\mathcal{B}$
        \STATE $\mathcal{L}_{\mathrm{MSE}} \leftarrow \frac{1}{|\mathcal{B}|}\sum_{(x_i, y_i)\in\mathcal{B}}\|\hat{y}_i - y_i\|_2^2$
        \STATE $\mathcal{L}_{\mathrm{straight}} \leftarrow \frac{1}{|\mathcal{B}|}\sum_{(x_i, y_i)\in\mathcal{B}}\sum_{j=1}^{L}\big\|\mathcal{F}_j(x_i)-\mathcal{F}_{j-1}(x_i)-\frac{1}{L}\{\mathcal{F}_L(x_i)-\mathcal{E}(x_i)\}\big\|_2^2$
        \STATE $\mathcal{L} \leftarrow \mathcal{L}_{\mathrm{MSE}} + \lambda\,\mathcal{L}_{\mathrm{straight}}$
        \STATE $\theta \leftarrow \theta - \eta\,\nabla_\theta\mathcal{L}$
    \ENDFOR
\ENDFOR
\RETURN $\theta$
\end{algorithmic}
\end{algorithm}

\begin{algorithm}[H]
\caption{Inference with $N_D$ dropped layers}
\label{alg:nn_inference}
\begin{algorithmic}[1]
\REQUIRE Trained parameters $\theta$, input $x$, number of layers to drop $0\le N_D\le L-1$
\ENSURE Predicted output $\hat{y}$
\STATE $a_0 \leftarrow \mathcal{E}(x)$
\FOR{$l = 1$ to $L-N_D$}
    \STATE $a_l \leftarrow \mathcal{S}_l(a_{l-1})$
\ENDFOR
\STATE $\Delta \leftarrow a_{L-N_D} - a_{L-N_D-1}$ \COMMENT{last computed update}
\STATE $\hat{a} \leftarrow a_{L-N_D} + N_D\,\Delta$
\RETURN $\hat{y} = \mathcal{D}(\hat{a})$
\end{algorithmic}
\end{algorithm}

\section{Experiments}

Most experiments compare models trained with and without the regularizer, and every task also includes a LayerDrop baseline. Model configurations are in \Cref{tab:model_configurations} in the appendix. We start with MNIST super-resolution with thorough experiments.

\begin{figure*}[!ht]
    \centering
    \begin{minipage}{.45\textwidth}
        \begin{tabular}{c|c|c|c|c }
         $\lambda$ & step loss & MSE & PSNR & SSIM\\
        \hline
        0 & 5.88e-1 & 1.91e-2 & 17.45 & 0.691 \\
        1e-5 & 4.33e-1 & 1.91e-2 & 17.45 & 0.689 \\
        1e-4 & 1.55e-1 & 1.91e-2 & 17.45 & 0.687 \\
        1e-3 & 4.01e-2 & 1.90e-2 & 17.47 & 0.709 \\
        1e-2 & 1.21e-2 & 1.89e-2 & 17.47 & 0.710 \\
        1e-1 & 2.85e-3 & 1.93e-2 & 17.47 & 0.730 \\
        1e0  & 5.12e-4 & 1.91e-2 & 17.49 &  0.724 \\
        1e+1 & 6.89e-5 & 1.87e-2 & 17.54 & 0.739 \\
        1e+2 & 8.92e-6 & 1.87e-2 & 17.54 & 0.729 \\
        1e+3 & 8.74e-7 & 1.88e-2 & 17.49 & 0.735 \\
        1e+4 & 9.46e-8 & 2.17e-2 & 16.88 & 0.677 \\
        1e+5 & 9.27e-9 & 2.51e-2 & 16.76 & 0.657 \\
        \hline
        \end{tabular}
        \captionof{table}{Full-depth performance of the MNIST super-resolution DiT for different $\lambda$.}
        \label{tab:result_varyingCoeff}
    \end{minipage}%
    \hfill
    \begin{minipage}{0.46\textwidth}
        \includegraphics[width=\linewidth]{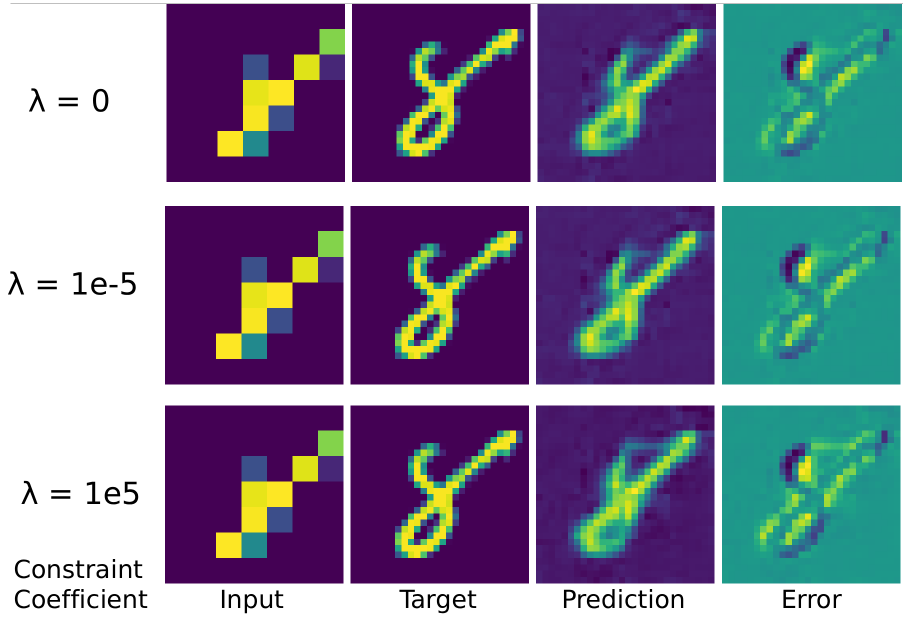}
        \caption{Visual results on one MNIST test digit for $\lambda=0$, $10^{-5}$ and $10^{5}$ (rows). Columns: the $7\times7$ input, the $28\times28$ target, the model prediction, and the error map. All three predictions reproduce the digit's shape.}
        \label{fig:viz_result_varyingCoeff}
    \end{minipage}
\end{figure*}

\begin{figure*}[!t]
    \centering
    \begin{minipage}{.49\textwidth}
        \includegraphics[width=\linewidth]{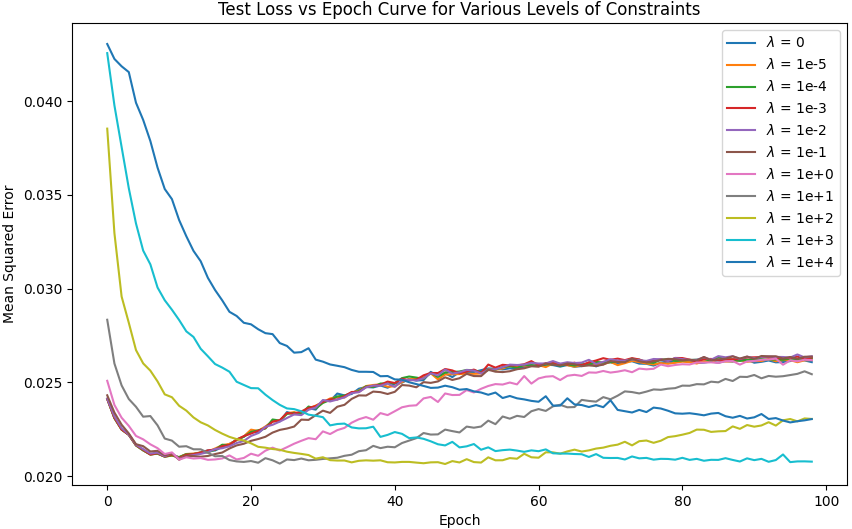}
        \caption{Test MSE per epoch on MNIST super-resolution for $\lambda$ from $0$ to $10^{4}$. Weakly regularized runs reach their minimum early and then rise; stronger constraints delay the minimum and flatten the later rise, and the $\lambda=10^4$ run is still falling at epoch 100.}
        \label{fig:MNIST_LossCurve_varyingCoeff}
    \end{minipage}%
    \hfill
    \begin{minipage}{0.49\textwidth}
        \includegraphics[width=\linewidth]{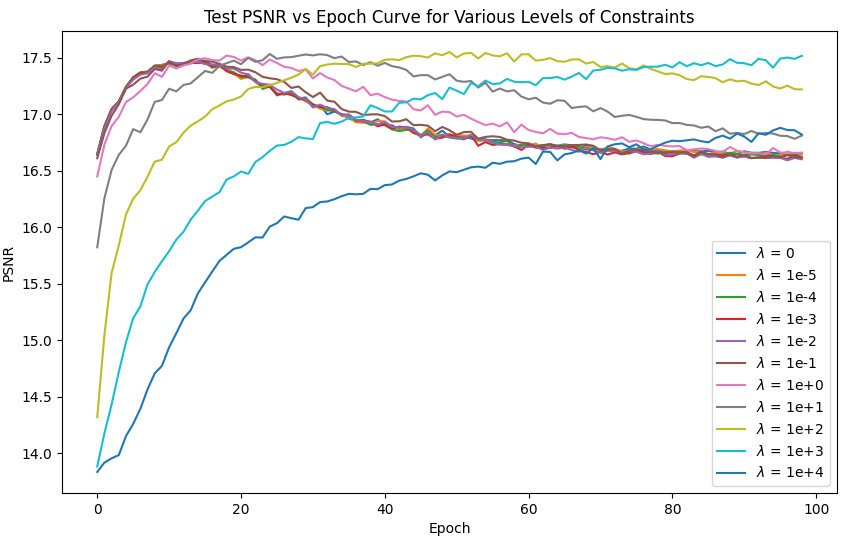}
        \caption{Test PSNR per epoch for the same runs as \Cref{fig:MNIST_LossCurve_varyingCoeff} (higher is better). Larger $\lambda$ climbs more slowly but peaks later and declines less after the peak; the $\lambda=10^4$ run has not peaked by epoch 100.}
        \label{fig:MNIST_PSNRCurve_varyingCoeff}
    \end{minipage}
\end{figure*}

\subsection{MNIST Super-Resolution}

A DiT \cite{peebles2023scalable} is trained on MNIST for $4\times$ super-resolution: the input is scaled down to $7\times7$ with nearest-neighbor interpolation, and the model recovers the original $28\times28$ image. We report the step loss (zero means all layers push the state in exactly the same direction and magnitude; lower means more aligned), and MSE, PSNR and SSIM against the target (lower MSE and higher PSNR and SSIM are better).

\paragraph{Latent dynamics and full-depth quality.}
The step loss falls from $5.88\times10^{-1}$ at $\lambda=0$ to $5.12\times10^{-4}$ at $\lambda=1$ and $9.27\times10^{-9}$ at $\lambda=10^5$ (\Cref{tab:result_varyingCoeff}). MSE and PSNR are essentially unchanged for $\lambda$ up to about $10^3$ (PSNR 17.45--17.54~dB). At $\lambda\ge10^4$ quality drops, to 16.88~dB at $10^4$ and 16.76~dB at $10^5$. Predictions look similar to the eye, see \Cref{fig:viz_result_varyingCoeff}.

\paragraph{Optimization.}
Stricter constraints slow optimization, but at moderate strengths the models reach similar minima (\Cref{fig:MNIST_LossCurve_varyingCoeff,fig:MNIST_PSNRCurve_varyingCoeff}). Both figures show test curves. For small $\lambda$ the test error is lowest after roughly 15--20 epochs and then rises, while larger $\lambda$ reaches its optimum later and degrades more slowly. The $\lambda=10^4$ run was stopped at 100 epochs while its curves were still improving. Very large $\lambda$ makes training unstable as well as slow, so for $\lambda\ge10^6$ we increase $\lambda$ on a logarithmic curriculum learning schedule.

\begin{table*}[t]
\centering
\begin{tabular}{c cc cc cc}
\toprule
& \multicolumn{2}{c}{$\lambda = 0$} & \multicolumn{2}{c}{$\lambda = 10^6$} & \multicolumn{2}{c}{LayerDrop} \\
\cmidrule(lr){2-3} \cmidrule(lr){4-5} \cmidrule(lr){6-7}
\textbf{\# Layers Dropped} & \textbf{PSNR} & \textbf{SSIM} & \textbf{PSNR} & \textbf{SSIM} & \textbf{PSNR} & \textbf{SSIM} \\
\midrule
0  & 17.52 & 0.692 & 16.40 & 0.631 & 17.56 & 0.693 \\
1  & 14.63 & 0.417 & 16.28 & 0.605 & 16.71 & 0.647 \\
2  & 14.21 & 0.394 & 16.03 & 0.569 & 15.88 & 0.580 \\
3  & 13.39 & 0.373 & 15.65 & 0.537 & 15.09 & 0.526 \\
4  & 13.09 & 0.389 & 15.18 & 0.508 & 14.11 & 0.461 \\
5  & 12.46 & 0.395 & 14.65 & 0.481 & 12.56 & 0.392 \\
6  & 11.46 & 0.382 & 14.09 & 0.455 & 12.10 & 0.375 \\
7  & 9.99  & 0.337 & 13.52 & 0.429 & 11.73 & 0.353 \\
8  & 7.57  & 0.265 & 12.94 & 0.403 & 10.21 & 0.302 \\
9  & 5.63  & 0.220 & 12.37 & 0.378 & 9.67  & 0.286 \\
10 & 4.17  & 0.193 & 11.81 & 0.353 & 7.32  & 0.211 \\
11 & 3.19  & 0.172 & 11.26 & 0.329 & 5.41  & 0.179 \\
\bottomrule
\end{tabular}
\caption{Layer dropping on MNIST super-resolution for models trained with $\lambda = 0$, $\lambda = 10^6$, and LayerDrop (each evaluated after removing the stated number of layers).}
\label{tab:layer_drops_vs_performance}
\end{table*}

\begin{table*}[t]
\center
\begin{tabular}{ c|c|c|c }
\toprule
 \# Layers Dropped & \# Parameters (M) & MSE (ours) & MSE (LayerDrop)\\
 \hline
 0 & 129.76 & 1.26e-2 & 1.22e-2\\
 1 & 119.13 & 1.98e-2 & 1.63e-2\\
 2 & 108.50 & 2.27e-2 & 2.07e-2\\
 3 & 97.87  & 2.50e-2 & 3.31e-2\\
 11 & 12.84 & 3.13e-2 & 7.16e-1\\
 \bottomrule
\end{tabular}
\caption{Moving MNIST: test MSE and model size as layers are dropped at inference, for the DiT trained with $\lambda=1$ (ours) and a DiT trained with LayerDrop.}
\label{tab:case2}
\end{table*}

\begin{table*}[!t]
\centering
\begin{tabular}{c c c c c cc}
\toprule
 & & & Unconstrained & LayerDrop & \multicolumn{2}{c}{Ours ($\lambda=0.1$)} \\
\cmidrule(lr){4-4} \cmidrule(lr){5-5} \cmidrule(lr){6-7}
Dataset & Layers Dropped & Parameters (M) & Test Acc.\ (\%) & Test Acc.\ (\%) & Step loss & Test Acc.\ (\%) \\
\midrule
MNIST
 & 0 & 1.1996 & 99.06 & 98.96 & 0.0509 & 98.90 \\
 & 2 & 0.8031 & 78.61 & 90.56 & 0.0352 & 98.05 \\
 & 4 & 0.4065 & 47.22 & 67.49 & 0.0138 & 93.85 \\
\midrule
CIFAR100
 & 0 & 1.2173 & 42.26 & 43.46 & 0.4761 & 45.84 \\
 & 2 & 0.8207 & 37.89 & 41.01 & 0.2741 & 40.09 \\
 & 4 & 0.4242 & 11.92 & 18.36 & 0.0953 & 23.27 \\
\bottomrule
\end{tabular}
\caption{Regularized 6-layer ViTs under layer dropping, compared with unconstrained models trained without the regularizer ($\lambda=0$) and with LayerDrop. The unconstrained step loss at full depth is 61.6684 (MNIST) and 4.9891 (CIFAR100).}
\label{tab:mnist}
\end{table*}

\paragraph{Post-hoc depth reduction.}
Dropping a single layer from the unconstrained model lowers PSNR from 17.52 to 14.63~dB, whereas the $\lambda=10^6$ model drops only from 16.40 to 16.28~dB (\Cref{tab:layer_drops_vs_performance}). With 11 of 12 layers removed, the unconstrained model reaches 3.19~dB and the $\lambda=10^6$ model 11.26~dB (\Cref{tab:11_of12}; \Cref{fig:1e6_perfromance} plots the $\lambda=10^6$ curve). The $\lambda=10^6$ model is still below the unconstrained model's full-depth quality at every depth, so the benefit is relative to dropping layers from an unconstrained model.

LayerDrop is the better choice for very mild reductions. It matches the unconstrained model at full depth (17.56 versus 17.52~dB) and beats the $\lambda=10^6$ model with one layer dropped (16.71 versus 16.28~dB). From two dropped layers on, the $\lambda=10^6$ model is better in PSNR (16.03 versus 15.88~dB at two), and the gap widens to 11.26 versus 5.41~dB with 11 layers dropped (\Cref{tab:layer_drops_vs_performance}).

Intermediate $\lambda$ values do not improve on $\lambda=0$ uniformly or monotonically (\Cref{tab:Effect_of_Dropedd_Layer_on_Quality}). For example, at $\lambda=1$ dropping six layers is worse than at $\lambda=0$ (8.4~dB versus 11.5~dB), and at $\lambda=10^2$ dropping one layer is worse (12.4~dB versus 14.6~dB).

\paragraph{Pushing the step consistency loss below machine precision.}
For $\lambda$ above $10^5$ the step loss reaches the float32 precision limit. To probe this regime we trained a model with $\lambda=10^9$ for 300 epochs on a logarithmically increasing schedule. Its step loss reached $1.04\times10^{-16}$, and it produces nearly identical outputs at every depth (PSNR 14.16~dB at 0 and at 11 layers dropped, SSIM 0.441, MSE $3.99\times10^{-2}$ throughout; \Cref{tab:layer_drops_vs_performance_4e9}). Its full-depth PSNR is far below the $\lambda=0$ model's 17.52~dB, and constant performance across depths is also what one would see if the layers contributed little to the output. We therefore treat this as a limiting case that is consistent with the intended mechanism (aligned updates make depth irrelevant), not as a useful operating point.

\begin{table*}[t]
\centering
\begin{tabular}{cccccc}
\toprule
Model & Layers Dropped & Layers Used & Acc. (\%) & Macro-F1 & LayerDrop Acc.\ (\%) \\
\midrule
ViT (unconstrained) & 0 & 12 & 88.48 & 0.7925 & -- \\
\midrule
 & 0 & 12 & 84.29 & 0.7492 & 87.36 \\
 & 1 & 11 & 84.29 & 0.7492 & 85.82 \\
 & 2 & 10 & 83.32 & 0.7396 & 84.61 \\
 & 3 &  9 & 82.03 & 0.7272 & 82.45 \\
 & 4 &  8 & 81.67 & 0.7157 & 78.52 \\
ViT & 5 &  7 & 75.44 & 0.6457 & 67.31 \\
(ours) & 6 &  6 & 73.39 & 0.6264 & 63.54 \\
 & 7 &  5 & 74.82 & 0.6441 & 46.79 \\
 & 8 &  4 & 71.83 & 0.6201 & 37.31 \\
 & 9 &  3 & 67.69 & 0.5592 & 32.58 \\
 & 10 &  2 & 61.88 & 0.5241 & 27.93 \\
 & 11 &  1 & 73.48 & 0.5932 & 23.42 \\
\midrule
EfficientNet-B5 (reported) & 0 & 572 & 84.51 & - & - \\
ResNet-101 (reported)      & 0 & 101 & 83.96 & - & - \\
VGGNet-19 (reported)       & 0 &  19 & 82.30 & - & - \\
AlexNet (reported)         & 0 &   8 & 77.32 & - & - \\
\bottomrule
\end{tabular}
\caption{LSCIDMR-S test performance as layers are dropped at inference ($\lambda = 0.001$). The LayerDrop column reports accuracy for a ViT trained with LayerDrop~\cite{fan2019reducing} and pruned to the same number of layers. Rows marked ``reported'' are taken from \cite{bai2021lscidmr}.}
\label{tab:LSCIDMR-Slayer_drop_performance}
\end{table*}

\subsection{Moving MNIST Video Prediction}

We train a DiT with $\lambda=1$ on Moving MNIST. MSE rises from $1.26\times10^{-2}$ at full depth to $2.50\times10^{-2}$ with 3 layers dropped (about $2\times$) and to $3.13\times10^{-2}$ with 11 of 12 dropped (about $2.5\times$), while the parameters used at inference fall from 129.76M to 12.84M (\Cref{tab:case2}). LayerDrop is better for mild reductions (MSE $1.22\times10^{-2}$ versus $1.26\times10^{-2}$ at full depth, $1.63\times10^{-2}$ versus $1.98\times10^{-2}$ with one layer dropped, and $2.07\times10^{-2}$ versus $2.27\times10^{-2}$ with two), but with three layers dropped it is worse ($3.31\times10^{-2}$ versus $2.50\times10^{-2}$), and with 11 dropped it is far worse ($7.16\times10^{-1}$ versus $3.13\times10^{-2}$, about $23\times$). We do not report an unconstrained Moving MNIST model under depth reduction, so the benefit relative to no regularization is not measured here.

\subsection{MNIST and CIFAR100 Classification}

Classification results are more mixed than the generative ones. We train 6-layer ViTs on MNIST \cite{lecun2010mnist} and CIFAR100 \cite{krizhevsky2009learning} with $\lambda=0$ (unconstrained), with LayerDrop, and with $\lambda=0.1$, and compare the three under layer dropping (\Cref{tab:mnist}).

The regularizer cuts the step loss from 61.7 to 0.05 on MNIST and from 5.0 to 0.48 on CIFAR100 (\Cref{tab:mnist}). At full depth the regularized MNIST model is slightly below the unconstrained one (98.90\% versus 99.06\%), while the regularized CIFAR100 model is above it (45.84\% versus 42.26\%). Under depth reduction, MNIST accuracy falls to 98.05\% with 2 of 6 layers removed and to 93.85\% with 4 removed, versus 78.61\% and 47.22\% for the unconstrained model and 90.56\% and 67.49\% for LayerDrop. On CIFAR100 our accuracy falls to 40.09\% with 2 removed (unconstrained 37.89\%, LayerDrop 41.01\%) and to 23.27\% with 4 removed (11.92\% and 18.36\%), a substantial loss in absolute terms. LayerDrop is better than ours at two removed layers on CIFAR100 and, by 0.06 points, at full depth on MNIST; ours is better at the other reduced depths. The harder task tolerates moderate reductions but not aggressive ones.

\subsection{Satellite Cloud Image Classification}

We use LSCIDMR-S, the single-label subset of the LSCIDMR benchmark \cite{bai2021lscidmr}: 104,390 satellite images of $256\times256$ pixels across 11 weather categories. We create reproducible, class-stratified 80\%/20\% train/test splits with a fixed random seed, resize images to $128\times128$, and train a 12-layer ViT (\Cref{tab:model_configurations}) with $\lambda=0.001$. We report accuracy and macro-F1 on the test set (\Cref{tab:LSCIDMR-Slayer_drop_performance}).

The unconstrained ViT is the most accurate model at full depth (88.48\%), about 4 points above our regularized ViT (84.29\%). The regularized ViT is comparable to the reported EfficientNet-B5 (84.51\%) and ResNet-101 (83.96\%) results, although those were obtained on a different split and are only indicative. Dropping one layer leaves accuracy and macro-F1 unchanged, and dropping up to four costs about 2.6 points (81.67\%). The LayerDrop ViT is more accurate than ours at full depth (87.36\% versus 84.29\%) and with up to three layers dropped (82.45\% versus 82.03\% at three), but ours is better from four layers dropped (81.67\% versus 78.52\%) and by a wide margin with eight dropped (71.83\% versus 37.31\%). Further removal degrades our accuracy steadily, to 61.88\% with two layers left. The curve is not monotonic: the one-layer model (73.48\%) beats the two-layer model (61.88\%), and the six-layer model (73.39\%) is below the five-layer model (74.82\%). We have no explanation and treat this as open; multiple seeds would be needed to tell whether the effect is systematic. Overall, modest layer reduction is feasible, while aggressive pruning compromises classification performance.

\begin{figure}[t]
    \centering
    \includegraphics[width=0.5\linewidth]{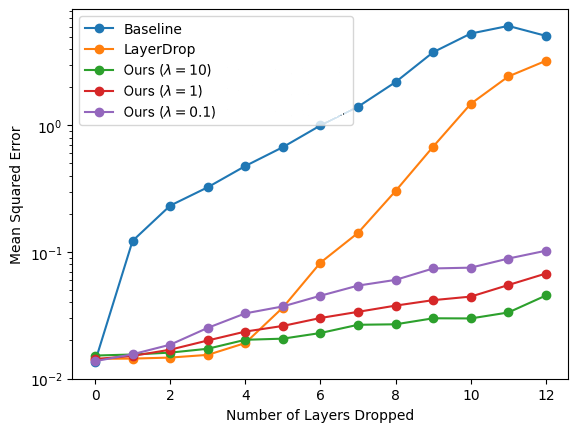}
    \caption{T2m test error (log scale) versus the number of blocks dropped from the 12-block DiTs. In the legend, \texttt{w} denotes $\lambda$ ($0.1$, $1$, $10$). The unconstrained baseline jumps by roughly an order of magnitude after a single block is removed. LayerDrop stays close to its full-depth error up to about four blocks and then rises steeply. The straightened models rise slowly and smoothly, and larger $\lambda$ gives lower error at large reductions.}
    \label{fig:t2m-depth}
\end{figure}

\begin{figure*}[!th]
    \centering
    \begin{minipage}{0.48\textwidth}
        \centering
        \includegraphics[width=\linewidth]{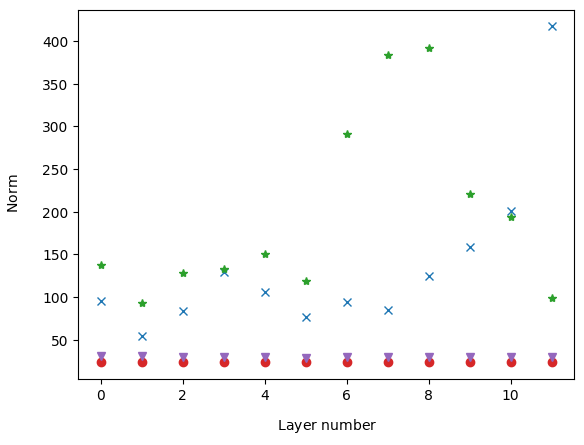}
    \end{minipage}\hfill
    \begin{minipage}{0.48\textwidth}
        \centering
        \includegraphics[width=\linewidth]{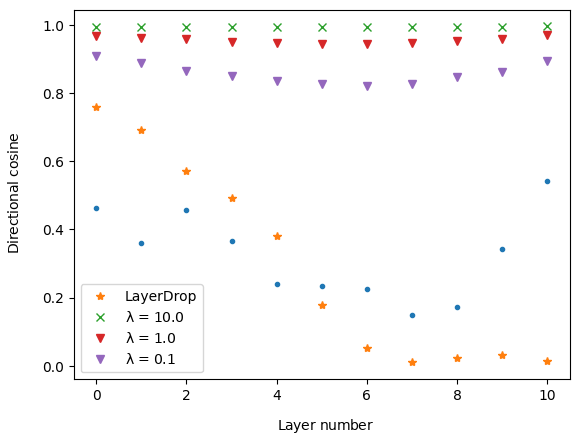}
    \end{minipage}
    \caption{Latent dynamics of the T2m models by block index. Left: norm of each block's update $\|x_{j+1}-x_j\|$. Right: mean cosine similarity between each block's update and the previous block's update (1 means perfectly aligned). The unconstrained baseline (blue) has large, uneven norms and weakly aligned updates. LayerDrop (orange) has large norms and updates that become almost orthogonal in later blocks. The straightened models ($\lambda=0.1$, $1$, $10$) have small, nearly constant norms and cosines close to one, ordered by $\lambda$.}
    \label{fig:t2m-dyn}

    \centering
    \includegraphics[width=\linewidth]{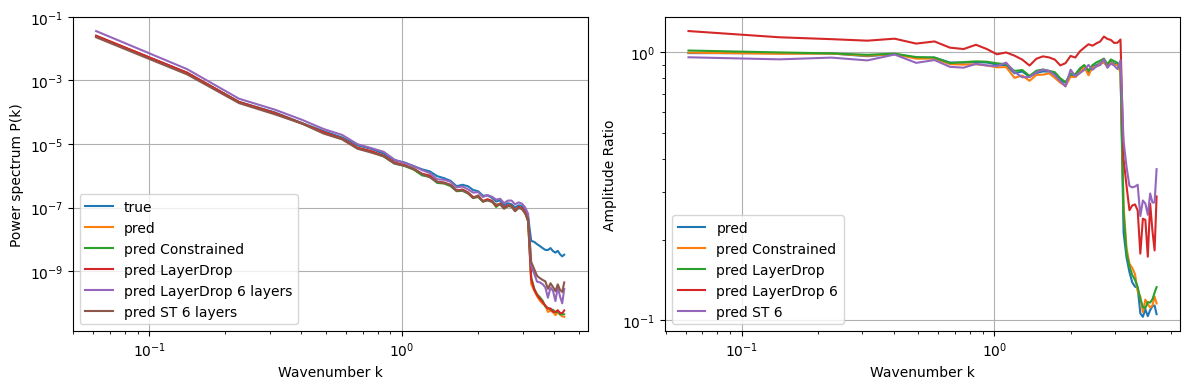}
    \caption{Left: radially averaged power spectra of the T2m target (``true'') and of predictions from the unconstrained, LayerDrop and straightened models, each at full depth and with six blocks removed (``ST'' = straightened). Right: ratio of predicted to target amplitude (1 means a perfect match). ``pred Constrained'' is the full-depth straightened model. All curves except LayerDrop with six blocks removed (ratio above one in the right panel) track the target up to a common cutoff at high wavenumber and fall off together beyond it. Removing six blocks does not change this behavior for the straightened model.}
    \label{fig:t2m-spectrum}
\end{figure*}

\subsection{2m Air Temperature Nowcasting}
\label{sec:t2m}

This task tests the method on a scientific problem and against an established depth-reduction method, LayerDrop \cite{fan2019reducing}. The task is 3-hour nowcasting of 2m temperature over the Indian subcontinent \cite{Rout_2025a,rout2026era5}. The data are 3-hourly temperature fields over South Asia spanning 40 years. Four consecutive fields (a 9-hour history) are mapped to the field 3 hours ahead, using $256\times256$ crops normalized as $(T-290)/10$, with the target hour as a 24-class conditioning label. We split in time (train through December 2020, test from January 2021) and report the mean squared error (in K$^2$) on the full test set over multiple seeds.

All models are 12-block DiTs: an unconstrained model, a LayerDrop model, and straightened models with $\lambda\in\{0.1,1,10\}$. At inference we keep the first $L-N_D$ blocks and apply either plain truncation or our extrapolation (\Cref{alg:nn_inference}) to every model, which separates the effect of training from that of the inference rule. At full depth all models perform similarly (MSE 1.39--1.54~K$^2$, i.e.\ RMSE about 1.18--1.24~K), so the regularizer costs little accuracy here, at most about 11\% higher MSE at $\lambda=10$ (\Cref{tab:t2m_loss_comparison}). Under depth reduction, the truncated unconstrained model degrades quickly. LayerDrop is the strongest baseline for small reductions, matching or beating the straightened models for $N_D\le4$. LayerDrop is therefore preferable for mild reductions and straightening for aggressive ones (\Cref{fig:t2m-depth}).

To check that updates are aligned and not merely small, we examine the update $x_{j+1}-x_j$ of each block $j$ for the unconstrained model, LayerDrop and the straightened models (\Cref{fig:t2m-dyn}). The left panel of \Cref{fig:t2m-dyn} shows the update norms. The unconstrained and LayerDrop models have large, uneven norms (roughly 50--400), whereas the straightened models keep a small, almost constant norm (roughly 25--30) across blocks. Small norms alone would not make a block safe to repeat, so the right panel reports the mean cosine similarity between each block's update and the previous block's update. For the straightened models it is close to one ($\approx1.0$ at $\lambda=10$, $\approx0.97$ at $\lambda=1$, $\approx0.82$--$0.91$ at $\lambda=0.1$), so consecutive updates point in nearly the same direction and repeating the last one is a reasonable approximation. For the unconstrained model it stays between about 0.15 and 0.55, and for LayerDrop it decays from about 0.75 to nearly zero in later blocks. Power spectra (\Cref{fig:t2m-spectrum}) show the full-depth models, and the straightened and LayerDrop models with six blocks removed, following the target up to a common cutoff (the unconstrained model with six blocks removed is not plotted), so depth reduction adds no extra smoothing at the resolved scales for these models. In the amplitude-ratio panel, the LayerDrop model with six blocks removed overestimates the target amplitude at low wavenumbers (ratio roughly 1.2--1.3), while the other curves stay near one. Beyond the cutoff every plotted model loses power relative to the target, with or without depth reduction.

\subsection{Computational Cost}

Removing layers reduces measured runtime by 51.5\% with 6 of 12 layers removed and by 88.9\% with 11 removed, relative to the full model (\Cref{tab:layer_drops_vs_runtime}). These savings come with the quality costs described above, and they exclude the added training cost of the regularizer and its slower convergence at large $\lambda$.

\begin{table*}[!t]
\centering
\begin{tabular}{c|cc}
\textbf{Layers Dropped} & \textbf{Runtime (m:ss)} & \textbf{Runtime Drop (\%)} \\
\hline
0  & 2:51 & 0.0 \\
1  & 2:40 & 6.4 \\
2  & 2:14 & 21.6 \\
3  & 2:02 & 28.7 \\
4  & 1:49 & 36.3 \\
5  & 1:36 & 43.9 \\
6  & 1:23 & 51.5 \\
7  & 1:10 & 59.1 \\
8  & 0:58 & 66.1 \\
9  & 0:44 & 74.3 \\
10 & 0:32 & 81.3 \\
11 & 0:19 & 88.9 \\
\hline
\end{tabular}
\vspace{0.25em}
\caption{Runtime as a function of the number of layers dropped.}
\label{tab:layer_drops_vs_runtime}

\end{table*}

\section{Discussion}

The regularizer reliably straightens updates: where we report the step loss it falls by one to several orders of magnitude ($5.9\times10^{-1}$ to $5.1\times10^{-4}$ at $\lambda=1$ on MNIST super-resolution, $61.7$ to $0.05$ on MNIST, and only $5.0$ to $0.48$, a factor of about 10, on CIFAR100); on T2m the cosine between consecutive updates rises from 0.15--0.55 to 0.82--1.0. Full-depth quality is preserved at moderate $\lambda$ on super-resolution but not uniformly elsewhere: the regularized MNIST model is slightly below its unconstrained counterpart, the CIFAR100 model is above it, and the LSCIDMR-S model is about 4 points below.

The clearest benefit is graceful degradation on MNIST super-resolution (3.19~dB versus 11.26~dB with 11 of 12 layers removed). The comparison with LayerDrop has the same shape on most tasks: LayerDrop is as good or better for mild reductions (one dropped layer on MNIST super-resolution, two on Moving MNIST and CIFAR100, three on LSCIDMR-S, four on T2m), and straightening is better for aggressive ones. MNIST classification is the exception, where straightening is already ahead at two dropped layers. We lack unconstrained baselines under depth reduction for Moving MNIST and LSCIDMR-S, so those results show costs, not benefits relative to no regularization. The $\lambda=10^9$ model confirms the mechanism (aligned updates make depth irrelevant) but is not a useful operating point.

\section{Limitations}

\begin{enumerate}
    \item \textbf{No broad baseline comparison.} Apart from LayerDrop (a structured-dropout method), we do not compare against post-hoc layer pruning, early exiting, or mixture-of-depths, so we cannot say how the approach compares in accuracy--cost trade-offs against them. The other comparisons are between models trained with and without the regularizer.
    \item \textbf{Small scale.} Experiments use MNIST, CIFAR100, Moving MNIST, one satellite dataset, and one nowcasting task, with models up to about 130M parameters. We have not tested larger language or vision--language models.
    \item \textbf{Task-dependent benefits.} CIFAR100 accuracy drops sharply when 4 of 6 layers are removed, the unconstrained model is more accurate on LSCIDMR-S, and LayerDrop is as good or better for mild reductions on most tasks.
    \item \textbf{Costs of large $\lambda$.} Very large $\lambda$ reduces full-depth quality, slows optimization, and requires a scheduled increase to train stably. The dependence on $\lambda$ is not monotonic at intermediate values (\Cref{tab:Effect_of_Dropedd_Layer_on_Quality}).
    \item \textbf{A proxy, without theory.} The step loss is a proxy for straightness. We give no theoretical analysis of when a straight trajectory is achievable or how the constraint changes the function class.
    \item \textbf{Manual depth selection.} Depth is chosen by the user at inference. We do not study input-dependent selection.
\end{enumerate}

\section{Conclusion}

We studied a trajectory-straightening regularizer for deep sequence models, motivated by viewing the layer chain as a discretized ODE. The regularizer penalizes misalignment of layer-wise updates, and a simple extrapolation rule replaces dropped layers at inference. Across a diffusion transformer and vision transformers, it reduced the directional-misalignment measure by one to several orders of magnitude and, at moderate strengths, left task quality largely unchanged on most tasks (LSCIDMR-S loses about 4 points). Models trained with it degraded much more gracefully under post-hoc depth reduction than unconstrained models, and than LayerDrop under aggressive reductions, although LayerDrop was as good or better for mild reductions: on MNIST super-resolution, removing 11 of 12 layers left 11.3~dB PSNR ($\lambda=10^6$) versus 3.2~dB without the constraint.

These gains have costs. Full-depth quality drops at large $\lambda$, accuracy on harder tasks such as CIFAR100 falls sharply under aggressive depth reduction, and optimization is slower. Our experiments also lack comparisons with early-exit, mixture-of-depths and post-hoc pruning methods, and with larger models; LayerDrop is the only baseline. Shaping the geometry of latent trajectories during training can make a single model usable at several depths. Whether this yields competitive accuracy--cost trade-offs remains to be shown.

Future work includes benchmarking against pruning, structured-dropout, and early-exit baselines; evaluating on larger models, including LLMs and VLMs; developing curvature-aware or data-dependent constraints; analyzing theoretically when a straight trajectory can be achieved; and combining the approach with input-dependent depth selection.




\bibliography{maine}

\begin{thebibliography}{56}
\providecommand{\natexlab}[1]{#1}
\providecommand{\url}[1]{\texttt{#1}}
\expandafter\ifx\csname urlstyle\endcsname\relax
  \providecommand{\doi}[1]{doi: #1}\else
  \providecommand{\doi}{doi: \begingroup \urlstyle{rm}\Url}\fi

\bibitem[Achiam et~al.(2023)Achiam, Adler, Agarwal, Ahmad, Akkaya, Aleman, Almeida, Altenschmidt, Altman, Anadkat, et~al.]{achiam2023gpt}
Josh Achiam, Steven Adler, Sandhini Agarwal, Lama Ahmad, Ilge Akkaya, Florencia~Leoni Aleman, Diogo Almeida, Janko Altenschmidt, Sam Altman, Shyamal Anadkat, et~al.
\newblock Gpt-4 technical report.
\newblock \emph{arXiv preprint arXiv:2303.08774}, 2023.

\bibitem[Arjovsky et~al.(2017)Arjovsky, Chintala, and Bottou]{arjovsky2017wasserstein}
Martin Arjovsky, Soumith Chintala, and L{\'e}on Bottou.
\newblock Wasserstein generative adversarial networks.
\newblock In \emph{International conference on machine learning}, pp.\  214--223. PMLR, 2017.

\bibitem[Bai et~al.(2021)Bai, Zhang, Zhang, Zheng, and Chen]{bai2021lscidmr}
Cong Bai, Minjing Zhang, Jinglin Zhang, Jianwei Zheng, and Shengyong Chen.
\newblock Lscidmr: Large-scale satellite cloud image database for meteorological research.
\newblock \emph{IEEE Transactions on Cybernetics}, 52\penalty0 (11):\penalty0 12538--12550, 2021.

\bibitem[Bai et~al.(2019)Bai, Kolter, and Koltun]{bai2019deep}
Shaojie Bai, J~Zico Kolter, and Vladlen Koltun.
\newblock Deep equilibrium models.
\newblock \emph{Advances in neural information processing systems}, 32, 2019.

\bibitem[Bansal et~al.(2018)Bansal, Chen, and Wang]{bansal2018can}
Nitin Bansal, Xiaohan Chen, and Zhangyang Wang.
\newblock Can we gain more from orthogonality regularizations in training deep networks?
\newblock \emph{Advances in Neural Information Processing Systems}, 31, 2018.

\bibitem[Bolukbasi et~al.(2017)Bolukbasi, Wang, Dekel, and Saligrama]{bolukbasi2017adaptive}
Tolga Bolukbasi, Joseph Wang, Ofer Dekel, and Venkatesh Saligrama.
\newblock Adaptive neural networks for efficient inference.
\newblock In \emph{International conference on machine learning}, pp.\  527--536. PMLR, 2017.

\bibitem[Chen et~al.(2018)Chen, Rubanova, Bettencourt, and Duvenaud]{chen2018neural}
Ricky~TQ Chen, Yulia Rubanova, Jesse Bettencourt, and David~K Duvenaud.
\newblock Neural ordinary differential equations.
\newblock \emph{Advances in neural information processing systems}, 31, 2018.

\bibitem[Daras et~al.(2024)Daras, Nie, Kreis, Dimakis, Mardani, Kovachki, and Vahdat]{daras2024warped}
Giannis Daras, Weili Nie, Karsten Kreis, Alex Dimakis, Morteza Mardani, Nikola Kovachki, and Arash Vahdat.
\newblock Warped diffusion: Solving video inverse problems with image diffusion models.
\newblock \emph{Advances in Neural Information Processing Systems}, 37:\penalty0 101116--101143, 2024.

\bibitem[Dehghani et~al.(2018)Dehghani, Gouws, Vinyals, Uszkoreit, and Kaiser]{dehghani2018universal}
Mostafa Dehghani, Stephan Gouws, Oriol Vinyals, Jakob Uszkoreit, and {\L}ukasz Kaiser.
\newblock Universal transformers.
\newblock \emph{arXiv preprint arXiv:1807.03819}, 2018.

\bibitem[Deng et~al.(2024)Deng, Wang, Zhang, Tai, and Tang]{deng2024dragvideo}
Yufan Deng, Ruida Wang, Yuhao Zhang, Yu-Wing Tai, and Chi-Keung Tang.
\newblock Dragvideo: Interactive drag-style video editing.
\newblock In \emph{European Conference on Computer Vision}, pp.\  183--199. Springer, 2024.

\bibitem[Devlin et~al.(2019)Devlin, Chang, Lee, and Toutanova]{devlin2019bert}
Jacob Devlin, Ming-Wei Chang, Kenton Lee, and Kristina Toutanova.
\newblock Bert: Pre-training of deep bidirectional transformers for language understanding.
\newblock In \emph{Proceedings of the 2019 conference of the North American chapter of the association for computational linguistics: human language technologies, volume 1 (long and short papers)}, pp.\  4171--4186, 2019.

\bibitem[Dosovitskiy(2020)]{dosovitskiy2020image}
Alexey Dosovitskiy.
\newblock An image is worth 16x16 words: Transformers for image recognition at scale.
\newblock \emph{arXiv preprint arXiv:2010.11929}, 2020.

\bibitem[Elbayad et~al.(2019)Elbayad, Gu, Grave, and Auli]{elbayad2019depth}
Maha Elbayad, Jiatao Gu, Edouard Grave, and Michael Auli.
\newblock Depth-adaptive transformer.
\newblock \emph{arXiv preprint arXiv:1910.10073}, 2019.

\bibitem[Fan et~al.(2019)Fan, Grave, and Joulin]{fan2019reducing}
Angela Fan, Edouard Grave, and Armand Joulin.
\newblock Reducing transformer depth on demand with structured dropout.
\newblock \emph{arXiv preprint arXiv:1909.11556}, 2019.

\bibitem[Finlay et~al.(2020)Finlay, Jacobsen, Nurbekyan, and Oberman]{finlay2020train}
Chris Finlay, J{\"o}rn-Henrik Jacobsen, Levon Nurbekyan, and Adam Oberman.
\newblock How to train your neural ode: the world of jacobian and kinetic regularization.
\newblock In \emph{International conference on machine learning}, pp.\  3154--3164. PMLR, 2020.

\bibitem[Ghosh et~al.(2024)Ghosh, Acharya, Saha, Jain, and Chadha]{ghosh2024exploring}
Akash Ghosh, Arkadeep Acharya, Sriparna Saha, Vinija Jain, and Aman Chadha.
\newblock Exploring the frontier of vision-language models: A survey of current methodologies and future directions.
\newblock \emph{arXiv preprint arXiv:2404.07214}, 2024.

\bibitem[Graves(2016)]{graves2016adaptive}
Alex Graves.
\newblock Adaptive computation time for recurrent neural networks.
\newblock \emph{arXiv preprint arXiv:1603.08983}, 2016.

\bibitem[Greff et~al.(2016)Greff, Srivastava, and Schmidhuber]{greff2016highway}
Klaus Greff, Rupesh~K Srivastava, and J{\"u}rgen Schmidhuber.
\newblock Highway and residual networks learn unrolled iterative estimation.
\newblock \emph{arXiv preprint arXiv:1612.07771}, 2016.

\bibitem[Gromov et~al.(2025)Gromov, Tirumala, Shapourian, Glorioso, and Roberts]{gromov2025unreasonable}
Andrey Gromov, Kushal Tirumala, Hassan Shapourian, Paolo Glorioso, and Daniel~A Roberts.
\newblock The unreasonable ineffectiveness of the deeper layers.
\newblock In \emph{International Conference on Learning Representations}, volume 2025, pp.\  81906--81920, 2025.

\bibitem[Gu \& Dao(2024)Gu and Dao]{gu2024mamba}
Albert Gu and Tri Dao.
\newblock Mamba: Linear-time sequence modeling with selective state spaces.
\newblock In \emph{First Conference on Language Modeling}, 2024.

\bibitem[Haber \& Ruthotto(2017)Haber and Ruthotto]{haber2017stable}
Eldad Haber and Lars Ruthotto.
\newblock Stable architectures for deep neural networks.
\newblock \emph{Inverse problems}, 34\penalty0 (1):\penalty0 014004, 2017.

\bibitem[Han et~al.(2021)Han, Huang, Song, Yang, Wang, and Wang]{han2021dynamic}
Yizeng Han, Gao Huang, Shiji Song, Le~Yang, Honghui Wang, and Yulin Wang.
\newblock Dynamic neural networks: A survey.
\newblock \emph{IEEE transactions on pattern analysis and machine intelligence}, 44\penalty0 (11):\penalty0 7436--7456, 2021.

\bibitem[H{\'e}naff et~al.(2019)H{\'e}naff, Goris, and Simoncelli]{henaff2019perceptual}
Olivier~J H{\'e}naff, Robbe~LT Goris, and Eero~P Simoncelli.
\newblock Perceptual straightening of natural videos.
\newblock \emph{Nature neuroscience}, 22\penalty0 (6):\penalty0 984--991, 2019.

\bibitem[Huang et~al.(2017)Huang, Chen, Li, Wu, Van Der~Maaten, and Weinberger]{huang2017multi}
Gao Huang, Danlu Chen, Tianhong Li, Felix Wu, Laurens Van Der~Maaten, and Kilian~Q Weinberger.
\newblock Multi-scale dense networks for resource efficient image classification.
\newblock \emph{arXiv preprint arXiv:1703.09844}, 2017.

\bibitem[Ilhan et~al.(2024)Ilhan, Chow, Hu, Huang, Tekin, Wei, Wu, Lee, Kompella, Latapie, et~al.]{ilhan2024adaptive}
Fatih Ilhan, Ka-Ho Chow, Sihao Hu, Tiansheng Huang, Selim Tekin, Wenqi Wei, Yanzhao Wu, Myungjin Lee, Ramana Kompella, Hugo Latapie, et~al.
\newblock Adaptive deep neural network inference optimization with eenet.
\newblock In \emph{2024 IEEE/CVF Winter Conference on Applications of Computer Vision (WACV)}, pp.\  1362--1371. IEEE, 2024.

\bibitem[Krizhevsky et~al.(2009)Krizhevsky, Hinton, et~al.]{krizhevsky2009learning}
Alex Krizhevsky, Geoffrey Hinton, et~al.
\newblock Learning multiple layers of features from tiny images.
\newblock 2009.

\bibitem[Kwon \& Ye(2025)Kwon and Ye]{kwon2025vision}
Taesung Kwon and Jong~Chul Ye.
\newblock Vision-xl: High definition video inverse problem solver using latent image diffusion models.
\newblock In \emph{Proceedings of the IEEE/CVF International Conference on Computer Vision}, pp.\  10465--10474, 2025.

\bibitem[Lan et~al.(2019)Lan, Chen, Goodman, Gimpel, Sharma, and Soricut]{lan2019albert}
Zhenzhong Lan, Mingda Chen, Sebastian Goodman, Kevin Gimpel, Piyush Sharma, and Radu Soricut.
\newblock Albert: A lite bert for self-supervised learning of language representations.
\newblock \emph{arXiv preprint arXiv:1909.11942}, 2019.

\bibitem[LeCun et~al.(2010)LeCun, Cortes, and Burges]{lecun2010mnist}
Yann LeCun, Corinna Cortes, and CJ~Burges.
\newblock Mnist handwritten digit database.
\newblock \emph{ATT Labs [Online]. Available: http://yann.lecun.com/exdb/mnist}, 2, 2010.

\bibitem[Li et~al.(2024)Li, Ma, Yang, and Yang]{li2024vidtome}
Xirui Li, Chao Ma, Xiaokang Yang, and Ming-Hsuan Yang.
\newblock Vidtome: Video token merging for zero-shot video editing.
\newblock In \emph{Proceedings of the IEEE/CVF Conference on Computer Vision and Pattern Recognition}, pp.\  7486--7495, 2024.

\bibitem[Liu et~al.(2025)Liu, Zhang, Huang, Xia, Wang, and Zhang]{liu2025vision}
Xiao Liu, Chenxu Zhang, Fuxiang Huang, Shuyin Xia, Guoyin Wang, and Lei Zhang.
\newblock Vision mamba: A comprehensive survey and taxonomy.
\newblock \emph{IEEE Transactions on Neural Networks and Learning Systems}, 2025.

\bibitem[Liu et~al.(2022)Liu, Gong, and Liu]{liu2022flow}
Xingchao Liu, Chengyue Gong, and Qiang Liu.
\newblock Flow straight and fast: Learning to generate and transfer data with rectified flow.
\newblock \emph{arXiv preprint arXiv:2209.03003}, 2022.

\bibitem[Liu et~al.(2021{\natexlab{a}})Liu, Meng, Zhou, Chen, and Xu]{liu2021faster}
Yijin Liu, Fandong Meng, Jie Zhou, Yufeng Chen, and Jinan Xu.
\newblock Faster depth-adaptive transformers.
\newblock In \emph{Proceedings of the AAAI Conference on Artificial Intelligence}, volume~35, pp.\  13424--13432, 2021{\natexlab{a}}.

\bibitem[Liu et~al.(2021{\natexlab{b}})Liu, Lin, Cao, Hu, Wei, Zhang, Lin, and Guo]{liu2021swin}
Ze~Liu, Yutong Lin, Yue Cao, Han Hu, Yixuan Wei, Zheng Zhang, Stephen Lin, and Baining Guo.
\newblock Swin transformer: Hierarchical vision transformer using shifted windows.
\newblock In \emph{Proceedings of the IEEE/CVF international conference on computer vision}, pp.\  10012--10022, 2021{\natexlab{b}}.

\bibitem[Men et~al.(2025)Men, Xu, Zhang, Yuan, Wang, Lin, Lu, Han, and Chen]{men2025shortgpt}
Xin Men, Mingyu Xu, Qingyu Zhang, Qianhao Yuan, Bingning Wang, Hongyu Lin, Yaojie Lu, Xianpei Han, and Weipeng Chen.
\newblock Shortgpt: Layers in large language models are more redundant than you expect.
\newblock In \emph{Findings of the Association for Computational Linguistics: ACL 2025}, pp.\  20192--20204, 2025.

\bibitem[Peebles \& Xie(2023)Peebles and Xie]{peebles2023scalable}
William Peebles and Saining Xie.
\newblock Scalable diffusion models with transformers.
\newblock In \emph{Proceedings of the IEEE/CVF international conference on computer vision}, pp.\  4195--4205, 2023.

\bibitem[Qian et~al.(2024)Qian, Dong, Zhang, Zang, Ding, Lin, and Wang]{qian2024streaming}
Rui Qian, Xiaoyi Dong, Pan Zhang, Yuhang Zang, Shuangrui Ding, Dahua Lin, and Jiaqi Wang.
\newblock Streaming long video understanding with large language models.
\newblock \emph{Advances in Neural Information Processing Systems}, 37:\penalty0 119336--119360, 2024.

\bibitem[Raposo et~al.(2024)Raposo, Ritter, Richards, Lillicrap, Humphreys, and Santoro]{raposo2024mixture}
David Raposo, Sam Ritter, Blake Richards, Timothy Lillicrap, Peter~Conway Humphreys, and Adam Santoro.
\newblock Mixture-of-depths: Dynamically allocating compute in transformer-based language models.
\newblock \emph{arXiv preprint arXiv:2404.02258}, 2024.

\bibitem[Rifai et~al.(2011)Rifai, Vincent, Muller, Glorot, and Bengio]{rifai2011contractive}
Salah Rifai, Pascal Vincent, Xavier Muller, Xavier Glorot, and Yoshua Bengio.
\newblock Contractive auto-encoders: Explicit invariance during feature extraction.
\newblock In \emph{Proceedings of the 28th international conference on international conference on machine learning}, pp.\  833--840, 2011.

\bibitem[Rout(2025)]{Rout_2025a}
Siddharth Rout.
\newblock Indian subcontinent 2m temperature field, Jul 2025.
\newblock URL \url{https://www.kaggle.com/datasets/siddharthrout/indian-subcontinent-temperature-field}.

\bibitem[Rout(2026)]{rout2026era5}
Siddharth Rout.
\newblock Era5-derived data and resources for generative modeling of 2-m air temperature over south asia.
\newblock 2026.

\bibitem[Rubanova et~al.(2019)Rubanova, Chen, and Duvenaud]{rubanova2019latent}
Yulia Rubanova, Ricky~TQ Chen, and David~K Duvenaud.
\newblock Latent ordinary differential equations for irregularly-sampled time series.
\newblock \emph{Advances in neural information processing systems}, 32, 2019.

\bibitem[Shi et~al.(2017)Shi, Gao, Lausen, Wang, Yeung, Wong, and Woo]{shi2017deep}
Xingjian Shi, Zhihan Gao, Leonard Lausen, Hao Wang, Dit-Yan Yeung, Wai-kin Wong, and Wang-chun Woo.
\newblock Deep learning for precipitation nowcasting: A benchmark and a new model.
\newblock \emph{Advances in neural information processing systems}, 30, 2017.

\bibitem[Shu et~al.(2025)Shu, Liu, Zhang, Qin, Zhou, Liang, Huang, and Zhao]{shu2025video}
Yan Shu, Zheng Liu, Peitian Zhang, Minghao Qin, Junjie Zhou, Zhengyang Liang, Tiejun Huang, and Bo~Zhao.
\newblock Video-xl: Extra-long vision language model for hour-scale video understanding.
\newblock In \emph{Proceedings of the Computer Vision and Pattern Recognition Conference}, pp.\  26160--26169, 2025.

\bibitem[Sun et~al.(2025)Sun, Pickett, Nain, and Jones]{sun2025transformer}
Qi~Sun, Marc Pickett, Aakash~Kumar Nain, and Llion Jones.
\newblock Transformer layers as painters.
\newblock In \emph{Proceedings of the AAAI Conference on Artificial Intelligence}, volume~39, pp.\  25219--25227, 2025.

\bibitem[Surat et~al.(2016)Surat, McDanel, and Kung]{teerapittayanon2016branchynet}
Teerapittayanon Surat, Bradley McDanel, and Hsiang-Tsung Kung.
\newblock Branchynet: Fast inference via early exiting from deep neural networks.
\newblock In \emph{2016 23rd international conference on pattern recognition (ICPR)}, pp.\  2464--2469. IEEE, 2016.

\bibitem[Tian et~al.(2024)Tian, Yang, Yang, Gao, Deng, Chen, Wang, Yu, Tao, Wan, et~al.]{tian2024videotetris}
Ye~Tian, Ling Yang, Haotian Yang, Yuan Gao, Yufan Deng, Jingmin Chen, Xintao Wang, Zhaochen Yu, Xin Tao, Pengfei Wan, et~al.
\newblock Videotetris: Towards compositional text-to-video generation.
\newblock \emph{Advances in Neural Information Processing Systems}, 37:\penalty0 29489--29513, 2024.

\bibitem[Touvron et~al.(2023)Touvron, Lavril, Izacard, Martinet, Lachaux, Lacroix, Rozi{\`e}re, Goyal, Hambro, Azhar, et~al.]{touvron2023llama}
Hugo Touvron, Thibaut Lavril, Gautier Izacard, Xavier Martinet, Marie-Anne Lachaux, Timoth{\'e}e Lacroix, Baptiste Rozi{\`e}re, Naman Goyal, Eric Hambro, Faisal Azhar, et~al.
\newblock Llama: Open and efficient foundation language models.
\newblock \emph{arXiv preprint arXiv:2302.13971}, 2023.

\bibitem[Vaswani et~al.(2017)Vaswani, Shazeer, Parmar, Uszkoreit, Jones, Gomez, Kaiser, and Polosukhin]{vaswani2017attention}
Ashish Vaswani, Noam Shazeer, Niki Parmar, Jakob Uszkoreit, Llion Jones, Aidan~N Gomez, {\L}ukasz Kaiser, and Illia Polosukhin.
\newblock Attention is all you need.
\newblock \emph{Advances in neural information processing systems}, 30, 2017.

\bibitem[Veit \& Belongie(2018)Veit and Belongie]{veit2018convolutional}
Andreas Veit and Serge Belongie.
\newblock Convolutional networks with adaptive inference graphs.
\newblock In \emph{Proceedings of the European conference on computer vision (ECCV)}, pp.\  3--18, 2018.

\bibitem[Veit et~al.(2016)Veit, Wilber, and Belongie]{veit2016residual}
Andreas Veit, Michael~J Wilber, and Serge Belongie.
\newblock Residual networks behave like ensembles of relatively shallow networks.
\newblock \emph{Advances in neural information processing systems}, 29, 2016.

\bibitem[Wang et~al.(2018)Wang, Yu, Dou, Darrell, and Gonzalez]{wang2018skipnet}
Xin Wang, Fisher Yu, Zi-Yi Dou, Trevor Darrell, and Joseph~E Gonzalez.
\newblock Skipnet: Learning dynamic routing in convolutional networks.
\newblock In \emph{Proceedings of the European conference on computer vision (ECCV)}, pp.\  409--424, 2018.

\bibitem[Wu et~al.(2021)Wu, Yao, Wang, and Long]{wu2021motionrnn}
Haixu Wu, Zhiyu Yao, Jianmin Wang, and Mingsheng Long.
\newblock Motionrnn: A flexible model for video prediction with spacetime-varying motions.
\newblock In \emph{Proceedings of the IEEE/CVF conference on computer vision and pattern recognition}, pp.\  15435--15444, 2021.

\bibitem[Yoshida \& Miyato(2017)Yoshida and Miyato]{yoshida2017spectral}
Yuichi Yoshida and Takeru Miyato.
\newblock Spectral norm regularization for improving the generalizability of deep learning.
\newblock \emph{arXiv preprint arXiv:1705.10941}, 2017.

\bibitem[Zheng et~al.(2024)Zheng, Peng, Yang, Shen, Li, Liu, Zhou, Li, and You]{zheng2024open}
Zangwei Zheng, Xiangyu Peng, Tianji Yang, Chenhui Shen, Shenggui Li, Hongxin Liu, Yukun Zhou, Tianyi Li, and Yang You.
\newblock Open-sora: Democratizing efficient video production for all.
\newblock \emph{arXiv preprint arXiv:2412.20404}, 2024.

\bibitem[Zhu et~al.(2024)Zhu, Liao, Zhang, Wang, Liu, and Wang]{zhu2024vision}
Lianghui Zhu, Bencheng Liao, Qian Zhang, Xinlong Wang, Wenyu Liu, and Xinggang Wang.
\newblock Vision mamba: Efficient visual representation learning with bidirectional state space model.
\newblock \emph{arXiv preprint arXiv:2401.09417}, 2024.

\end{thebibliography}
\bibliographystyle{tmlr}

\appendix
\begin{table*}[h]
\textbf{APPENDIX\vspace{1cm}}
\centering
\small
\begin{tabular}{lcccccc}
\hline
Variable & MNIST & CIFAR100 & MNIST (SR) & Moving MNIST & LSCIDMR-S & T2m \\
\hline
Task & Classification & Classification & Super-resolution & Video prediction & Classification & Nowcasting \\
Architecture & ViT & ViT & DiT & DiT & ViT & DiT\\
Image Size & 28 & 32 & 28 & 32 & 128 & 256\\
\# Input Channels & 1 & 3 & 1 & 5 & 3 & 4\\
\# Output Channels & 10 & 100 & 1 & 5 & 11 & 1\\
Hidden Size & 128 & 128 & 768 & 768 & 512 & 384\\
\# Layers & 6 & 6 & 12 & 12 & 12 & 12\\
Patch Size & 7 & 4 & 7 & 2 & 32 & 16\\
\# Attention Heads & 4 & 4 & 8 & 8 & 4 & 4\\
\hline
\end{tabular}
\caption{Model configurations used for MNIST and CIFAR-100 image classification, MNIST super-resolution, Moving MNIST video prediction, LSCIDMR-S cloud classification, and T2m nowcasting.}
\label{tab:model_configurations}
\end{table*}


\begin{table*}[!ht]

\centering
\begin{tabular}{c|ccc}
$\lambda$ & MSE & PSNR & SSIM  \\
\hline
0 & 4.8e-1 & 3.19 & 0.17  \\
1e-4 & 4.8e-1 & 3.20 & 0.17  \\
1e-2 & 4.9e-1 & 3.12 & 0.17\\
1e-1 & 4.4e-1 & 3.59 & 0.17 \\ 
1e+0 & 1.9e-1 & 7.18 & 0.22 \\
1e+1 & 1.0e-1 & 9.77 & 0.28 \\
1e+2 & 1.1e-1 & 9.48 & 0.28 \\
1e+4 & 1.1e-1 & 9.46 & 0.30 \\
1e+5 & 8.9e-2 & 10.6 & 0.32 \\ 
1e+6 & 8.2e-2 & 11.3 & 0.33 \\ 
\hline
\end{tabular}
\vspace{1.2em}
\caption{Performance for different constraint coefficients after dropping 11 out of 12 layers.}
\label{tab:11_of12}
\end{table*}

\begin{table*}[!th]
\centering
\begin{tabular}{c|lll|lll|lll}
       & \multicolumn{3}{c|}{Original}                       & \multicolumn{3}{c|}{Dropping 1 layer}                       & \multicolumn{3}{c}{Dropping 6 layers}                                             \\ \cline{2-10}
       
$\lambda$ & \multicolumn{1}{l|}{MSE} & \multicolumn{1}{l|}{PSNR} & SSIM & \multicolumn{1}{l|}{MSE} & \multicolumn{1}{l|}{PSNR} & SSIM & \multicolumn{1}{l|}{MSE} & \multicolumn{1}{l|}{PSNR} & SSIM  \\ \hline

0      & \multicolumn{1}{l|}{1.9e-2}   & \multicolumn{1}{l|}{17.5}    & 0.69    & \multicolumn{1}{l|}{3.4e-2}   & \multicolumn{1}{l|}{14.6}    & 0.41    & \multicolumn{1}{l|}{7.3e-2}   & \multicolumn{1}{l|}{11.5}    & 0.38        \\ 


1e-4       & \multicolumn{1}{l|}{1.9e-2}    & \multicolumn{1}{l|}{17.5}     &   0.69   & \multicolumn{1}{l|}{3.2e-2}    & \multicolumn{1}{l|}{15.2}     &   0.46   & \multicolumn{1}{l|}{7.3e-2}    & \multicolumn{1}{l|}{11.5}     &    0.39      \\ 


1e-2       & \multicolumn{1}{l|}{1.9e-2}    & \multicolumn{1}{l|}{17.5}     &   0.71   & \multicolumn{1}{l|}{2.6e-2}    & \multicolumn{1}{l|}{16.0}     &  0.60   & \multicolumn{1}{l|}{7.6e-2}    & \multicolumn{1}{l|}{11.3}     &    0.36      \\ 
  

1e+0       & \multicolumn{1}{l|}{1.9e-2}    & \multicolumn{1}{l|}{17.5}     &   0.72   & \multicolumn{1}{l|}{3.1e-2}    & \multicolumn{1}{l|}{15.2}     &   0.50   & \multicolumn{1}{l|}{1.5e-1}    & \multicolumn{1}{l|}{8.4}     &   0.28      \\ 


1e+2       & \multicolumn{1}{l|}{1.9e-2}    & \multicolumn{1}{l|}{17.5}     &   0.73   & \multicolumn{1}{l|}{5.6e-2}    & \multicolumn{1}{l|}{12.4}     &  0.45   & \multicolumn{1}{l|}{1.1e-1}    & \multicolumn{1}{l|}{9.5}     &     0.31    \\ 
  

1e+4       & \multicolumn{1}{l|}{2.2e-2}    & \multicolumn{1}{l|}{16.9}     &   0.68   & \multicolumn{1}{l|}{2.3e-2}    & \multicolumn{1}{l|}{16.5}     &   0.59   & \multicolumn{1}{l|}{5.3e-2}    & \multicolumn{1}{l|}{12.8}     &    0.42     \\ 
  
1e+5       & \multicolumn{1}{l|}{2.5e-2}    & \multicolumn{1}{l|}{16.8}     &  0.66    & \multicolumn{1}{l|}{2.4e-2}    & \multicolumn{1}{l|}{16.5}     &   0.62   & \multicolumn{1}{l|}{4.4e-2}    & \multicolumn{1}{l|}{13.7}     &    0.45     \\ \hline
\end{tabular}
\vspace{0.5em}
\caption{Performance of the model with various constraint coefficients, $\lambda$s, for the MNIST super-resolution DiT, at full depth and after dropping 1 and 6 of 12 layers.}
\label{tab:Effect_of_Dropedd_Layer_on_Quality} 
\end{table*}

\begin{table*}
\centering
\begin{tabular}{c|ccc}
\textbf{\# of Layers Dropped} & \textbf{MSE} & \textbf{PSNR} & \textbf{SSIM} \\
\hline
0  & 3.988e-02 & 14.161 & 0.441 \\
2  & 3.988e-02 & 14.162 & 0.441 \\
4  & 3.989e-02 & 14.161 & 0.441 \\ 
6  & 3.989e-02 & 14.161 & 0.441 \\
8  & 3.989e-02 & 14.161 & 0.441 \\
10 & 3.989e-02 & 14.162 & 0.441 \\
11 & 3.989e-02 & 14.162 & 0.441 \\
\hline
\end{tabular}
\vspace{0.25em}
\caption{Effect on performance of prediction for model trained with $\lambda = 1e9$, that pushes step loss below 32-bit precision limit, after dropping numerous hidden layers.}
\label{tab:layer_drops_vs_performance_4e9}
\end{table*}

\begin{table*}[t]
\centering
\begin{tabular}{lc}
\hline
\textbf{Model} & \textbf{MSE (K$^2$)} \\
\hline
Baseline     & 1.3855 \\
LayerDrop  & 1.4498 \\
Ours, $\lambda=0.1$ & 1.3951 \\
Ours, $\lambda=1$  & 1.4535 \\
Ours, $\lambda=10$    & 1.5420 \\
\hline
\end{tabular}
\caption{Full-depth test MSE on T2m (K$^2$).}
\label{tab:t2m_loss_comparison}
\end{table*}

\begin{figure*}[t]
    \centering
\begin{minipage}{0.48\textwidth}
\includegraphics[width=\linewidth]{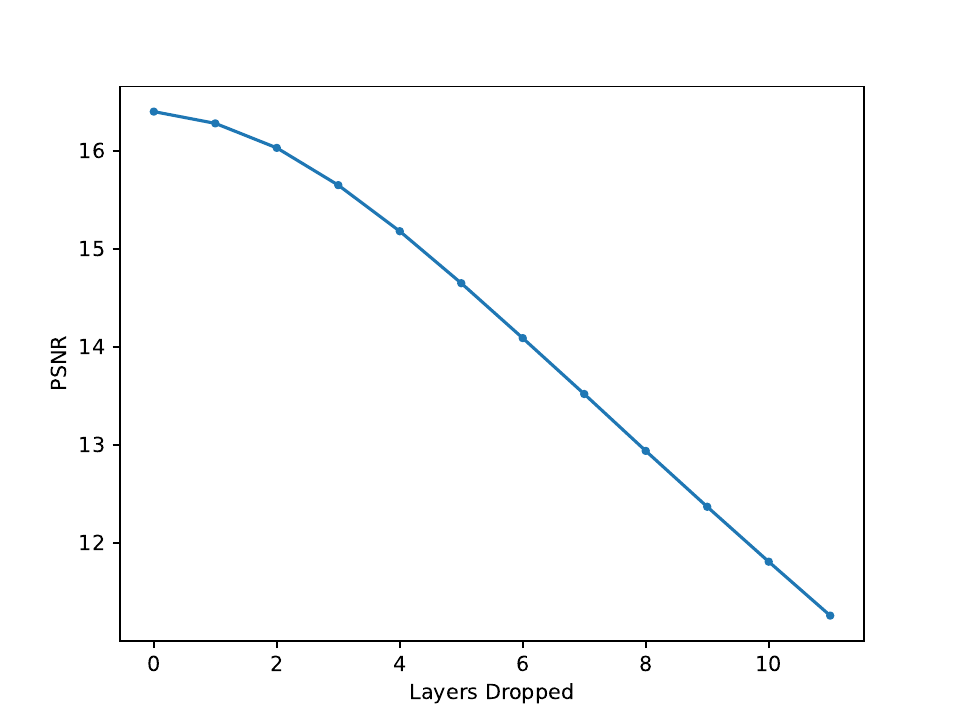}
\end{minipage}
\hfill
\begin{minipage}{0.48\textwidth}
\includegraphics[width=\linewidth]{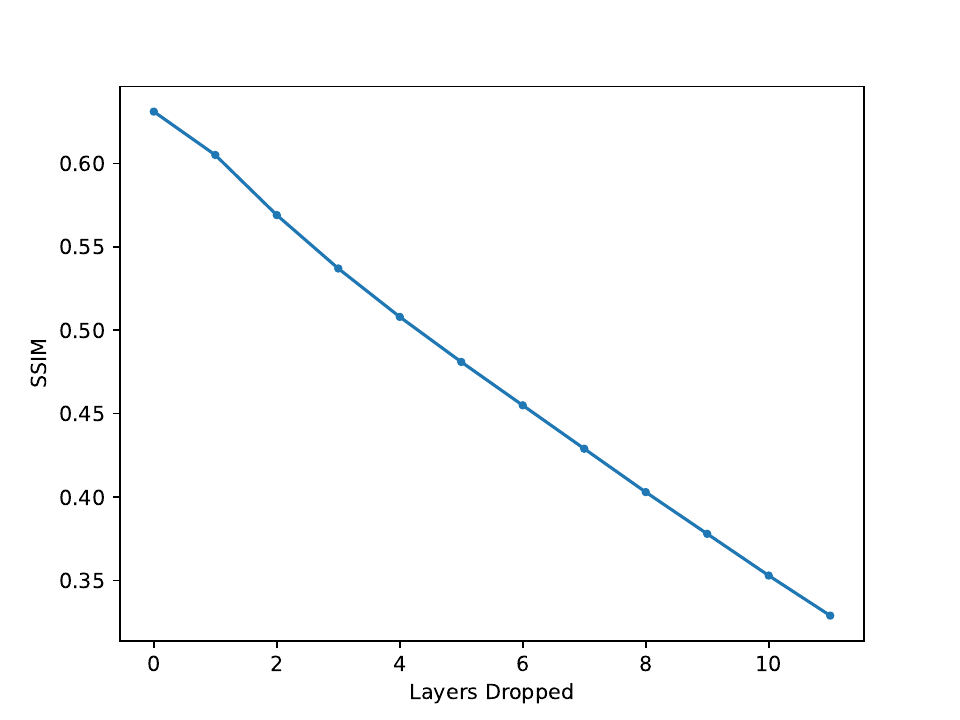}
\end{minipage}
    \caption{MNIST super-resolution with $\lambda=10^6$: PSNR (left) and SSIM (right) as 0 to 11 of 12 layers are dropped (values in \Cref{tab:layer_drops_vs_performance}). PSNR falls from 16.4 to 11.3~dB and SSIM from 0.63 to 0.33, steadily and without a sharp drop, unlike the unconstrained model, which loses about 2.9~dB after the first layer (\Cref{tab:layer_drops_vs_performance}).}
    \label{fig:1e6_perfromance}
\end{figure*}

\end{document}